\documentclass[conference]{IEEEtran}

\IEEEoverridecommandlockouts
\IEEEpubid{\makebox[\columnwidth]{978-1-6654-XXXX-X/24/\$00.00~\copyright~2025 IEEE \hfill}
\hspace{\columnsep}\makebox[\columnwidth]{ }}

\usepackage{graphicx}      
\usepackage{cite}
\usepackage{amsmath}
\usepackage{color,soul}

\usepackage{caption}
\usepackage{subcaption}
\usepackage{verbatim}
\usepackage{algorithm}
\usepackage{algpseudocode}
\usepackage{amsmath}
\usepackage{tikz}
\usepackage{float}

\begin{document}

\title{Blockchain Federated Learning for Sustainable Retail: Reducing Waste through Collaborative Demand Forecasting} 



\author{
    \IEEEauthorblockN{
        Fabio Turazza\IEEEauthorrefmark{1}, 
        Alessandro Neri\IEEEauthorrefmark{1}\IEEEauthorrefmark{3}, 
        Marcello Pietri\IEEEauthorrefmark{1}, 
        Maria Angela Butturi\IEEEauthorrefmark{1}, 
        Marco Picone\IEEEauthorrefmark{1}
        Marco Mamei\IEEEauthorrefmark{1}\IEEEauthorrefmark{2}
    }
    \IEEEauthorblockA{\IEEEauthorrefmark{1}DISMI, University of Modena and Reggio Emilia, Italy}
    \IEEEauthorblockA{\IEEEauthorrefmark{2}En\&Tech, University of Modena and Reggio Emilia, Italy}
    \IEEEauthorblockA{\IEEEauthorrefmark{3}DIN, University of Bologna, Italy}
    \IEEEauthorblockA{Email: name.surname@unimore.it}
}

\maketitle

\begin{abstract}                
Effective demand forecasting is crucial for reducing food waste. However, data privacy concerns often hinder collaboration among retailers, limiting the potential for improved predictive accuracy. In this study, we explore the application of Federated Learning (FL) in Sustainable Supply Chain Management (SSCM), with a focus on the grocery retail sector dealing with perishable goods. We develop a baseline predictive model for demand forecasting and waste assessment in an isolated retailer scenario. Subsequently, we introduce a Blockchain-based FL model, trained collaboratively across multiple retailers without direct data sharing. Our preliminary results show that FL models have performance almost equivalent to the ideal setting in which parties share data with each other, and are notably superior to models built by individual parties without sharing data, cutting waste and boosting efficiency.
\end{abstract}



\section{INTRODUCTION}

Food waste is increasingly recognised as an issue due to its high environmental, social, and economic impact. Food production accounts for 26\% of total greenhouse gas (GHG) emissions, and 26.7\% of the food is not even consumed \cite{Poore2018}. Thus, food waste from supply chain losses and end-consumers accounts for 6\% of total emissions.
Sustainable Development Goal 12.3 explicitly states to "halve per capita global food waste at the retail and consumer levels and reduce food losses along production and supply chains, including post-harvest losses". However, the food loss index is 98.3 with the 2021 food loss percentage at 13.23\% (unchanged from 2016). These results are deeply influenced by high-income countries, which tend to waste more food \cite{Narvanen2023}.
Almost two thirds of all food waste in Europe is the result of the consumer sector, including retail. Retail plays an important role in minimising waste by continually balancing the need of high stocks (to guard against risky stockouts) and lean inventories (to limit overstock effects) \cite{Riesenegger2022}.
Re-engineering internal processes via improved stock management and demand forecasting is considered as a beneficial and efficient supply chain strategy to reduce food losses at the retail level \cite{Huang2021}.

The retail sector has seen a paradigm shift in values over the last decades. Corporate organisations are increasingly focusing on their long-term social impact alongside their short-term profits. Incorporating sustainability concepts into their strategies has become mandatory to align value chain goals \cite{Berning2015}. 
Sustainable development emphasizes an effort to improve the current scenario (environmental, economic, and social) without compromising future generations. Therefore, Sustainable Supply Chain Management (SSCM) integrates the triple bottom line in managing material, information, and capital flows \cite{Seuring2008}. Digital technologies (i.e., Internet-of-Things (IoT) and data analytics) enhance organizations with real-time data capabilities and facilitate simplified information sharing \cite{Kumar2023}. This capability allows to extract value from data, forecasting demand rather than other more labor-intensive strategies. 

In this context, the work aims at proposing an efficient approach for integrating IoT and Federated Learning (FL) in the retail supply chain of perishable products. Food shops can avoid over-purchasing and waste by leveraging advanced demand forecasting models. However, shops may have unreliable or sparse datasets. A feasible approach may involve leveraging information sharing among competitors to augment their own datasets. Nevertheless, retailers operating within the same sector might be reluctant to share sensitive data, but through FL, they can maintain data privacy and control. As a matter of fact, FL enables model parameters sharing without compromising private data, tapping into collective information and improving their predictive performance \cite{Zheng2023}.

We present an empirical case study to test the aforementioned hypothesis using a dataset, which we will refer to as the \textit{Walmart Database} \cite{Ramakrishnan2022}. Our case study represents a typical SSCM scenario. Actually, competing interests and data privacy concerns often limit collaboration. The objective we aim to achieve is to demonstrate the validity of a privacy-oriented Federated method based on Blockchain, comparing it with other basic approaches.

\section{BACKGROUND}
\label{background}

\begin{figure*}[h!]
\centering
A)\includegraphics[scale=0.7]{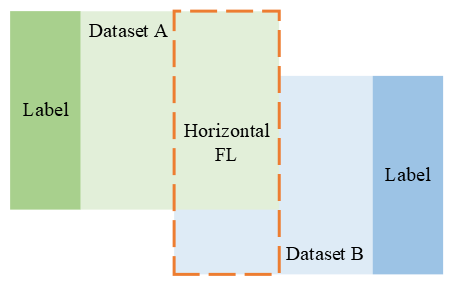}
B)\includegraphics[scale=0.7]{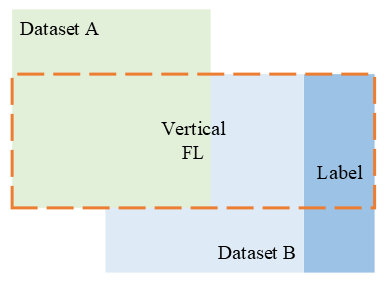}
C)\includegraphics[scale=0.7]{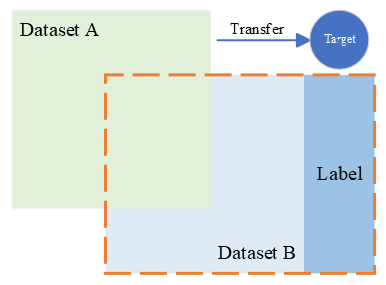}
\caption{Different categorizations of FL: {\bf A)} Horizontal FL, {\bf B)} Vertical FL, {\bf C)} Federated Transfer Learning}
\label{partition}
\end{figure*}

\begin{table*}[h]
    \centering
    \caption{Literature review comparison table}\label{comparison}
    \begin{tabular}{l c c c c l}
\hline
Ref. & Sector & FL Type & Sustainability Focus & Perishable Focus & Objective \\
\hline
\cite{Wang2022} & Retail & V & N & N & Demand Forecasting \\
\cite{Liu2022} & Automotive & V & N & N & Demand Forecasting \\
\cite{Li2021} & E-commerce & H & Y & N & Demand Forecasting \\
\cite{Durrant2022} & Agri-food & H & Y & Y & Yield Forecasting \\
\cite{Wan2023} & Closed-loop supply chain & N/A & Y & N & FL Impact Assessment  \\
\cite{Sun2023} & Energy & N/A & Y & N & Sustainable Production \\
This work & Grocery & V & Y & Y & Waste Reduction \\
\hline
\end{tabular}
\parbox{0.85\textwidth}{
    \footnotesize{Note: H - Horizontal, V - Vertical; Y - Yes, N - No}}
\end{table*}



Sustainable practices promotion within supply chains rely on cooperation. The implementation of IoT and data analytics could enhance supply chain transparency beyond corporate boundaries, which opens up possibilities for improvements \cite{Birkel2021}. However, the fear of transparency and large investments may discourage small and medium-sized businesses from disclosing information \cite{Dias2021}. In this context, the food industry still strives to adapt to customer demand. Yet, forecasting techniques (e.g., machine learning) boost efficiency and agility, but advanced machine learning algorithm exploitation is still at an early stage \cite{Rodrigues2024}. 

Traditional machine learning models merge different data-sets in a single one. This approach risks sensitive data leakage \cite{Zhang2022}. Therefore, FL, proposed by \cite{McMahan2016}, trains locally models without sharing raw data. As described in \cite{Wen2023} and \cite{Turazza2024MEDES}, FL takes into account $N$ participants, i.e. members that want to leverage trained model. Users merge their datasets $\{p_1, p_2, \dots, p_N\}$ to train the model. The FL model is trained by minimizing the loss function in Eq. \ref{fl_loss}. Where $n_k$ is the amount of data on the member $k$ and $F_k(w)$ is the local objective function. 
\begin{align}
    \min f(w) = \sum_{k=1}^N \frac{n_k}{n} F_k(w) \label{fl_loss}
\end{align}
The global model parameters are set up in a central server. Each client download the global parameters, trains its own model, and finally updates the global model round after round \cite{Zheng2023}. FL can be divided in three primary categorizations: horizontal FL, vertical FL, and Federated transfer learning. Horizontal type deals with overlapping features among the datasets of its members. In contrast, vertical one deals with overlapping users with less overlapping features. The last type rarely has overlapping datasets but it can overcome data scarcity by transferring learning \cite{Zhang2021}. Figure \ref{partition} shows different distribution patterns of sample space and feature space of data.

There are already examples in the literature regarding applications of FL in supply chain management. \cite{Wang2022} develop a vertical FL for the retail sector demand forecasting. This study integrates different datasets, such as social networks, e-commerce, and retail trader datasets. The paper shows that Fed-LSTM has better model's metrics (i.e., MSE, RMSE, and SD) against three other methods, LR, XGboost, and GBDT. \cite{Liu2022} still use vertical FL to perform a series of experiments to improve demand forecasting in the automotive retail sector. The study employees gradient descent methods and loss computation to improve the overall performance. 

In the context of SSCM, research on FL applications is still in its early stages. Nonetheless, sustainability topics have been explored in some studies. \cite{Li2021} addresses the challenges of e-commerce supply chains during public health emergencies. This work integrates horizontal FL and ConvLSTM to improve demand forecasting by better handling multi-dimensional time-series data. \cite{Durrant2022} propose a technical solution to overcome the challenges of data sharing in the agri-food sector, optimising production through soybean yield forecast. The model help enhancing food safety and net-zero targets. \cite{Wan2023} design a closed-loop supply chain entangling a manufacturer, a retailer, and a third party. The study underline the importance of information reliability and data security issues in coordination. \cite{Sun2023} integrate Blockchain and FL to manage energy demand and distribution. Lastly, \cite{Zheng2024} propose an adaptive FL model and test it using e-commerce platform data, suggesting that research in other contexts would be useful. This research propose a new approach that aims at reducing flaws in outages and improving sustainability, as demonstrated by the experimental analysis. 

Even though existing studies regarding FL often focus on peculiar aspects of sustainability through optimisation or forecasting, such as economic (i.e., improving revenues) and social (i.e., targeting privacy and security concerns), this work focuses on demand forecasting with the goal of waste prevention in supply chain management, aligning with the lean and green goals of SSCM. We aim to show the effectiveness of FL in contributing to perishable product waste reduction. Our approach demonstrates FL's versatility in addressing a variety of supply chain management challenges, particularly those related to sustainability.


\section{EXPERIMENTAL SETUP}
\label{exp}



The experiments for our study were conducted using the aforementioned \textit{Walmart Database} \cite{Ramakrishnan2022}. This dataset provides a comprehensive view of Walmart's operations, encompassing 45 stores selected for basic analysis from a total of 10,585 stores globally. Walmart utilizes its extensive big data database for various operational improvements. These include enhancing store checkout processes, efficiently managing steps in the supply chain, optimizing the product assortment offered in stores, and tailoring the shopping experience to individual customer preferences.

The dataset includes weekly sales data from 2010 to 2012 for these 45 stores, along with influencing factors like holidays, temperature, fuel price, Consumer Price Index (CPI), and unemployment rates. Notable holiday events like the Super Bowl, Labour Day, Thanksgiving, and Christmas are included, providing valuable insights into sales patterns during these periods.

\begin{itemize}
  \item \textit{Store}: Store Numbers ranging from 1 to 45.
  \item \textit{Date}: The Week of Sales.
  \item \textit{Weekly\_Sales}: The sales of the given store in the given week.
  \item \textit{Holiday\_Flag}: Indicates if the week includes a special holiday (1) or is a full working week (0).
  \item \textit{Temperature}: Average Temperature of the week.
  \item \textit{Fuel\_Price}: Price of the Fuel in the region.
  \item \textit{CPI}: Customer Price Index.
  \item \textit{Unemployment}: Unemployment rate of the region.
\end{itemize}

This dataset is vital for our case study, representing a typical SSCM scenario, where various factors influencing sales and supply chain dynamics can be analyzed. In Figure \ref{fig:shops45}, we present a graph showcasing the weekly sales data of the stores, with each line representing a different store. This visualization allows for a clear understanding of sales trends and variations across different Walmart locations. We address the challenge of waste reduction in perishable products' retail supply chains through FL. In this approach, we focus not only on mitigating over-purchasing and waste in food shops, but also on potential improvements and product transfers between various actors in the supply chain. This holistic view takes into account factors like fuel prices to enhance demand forecasting and supply chain efficiency. Results highlighted in Figure \ref{fig:fl_vs_all_vs_single} are obtained by comparing three different approaches:

\begin{figure}[htb]
\begin{center}
\includegraphics[width=8.8cm]{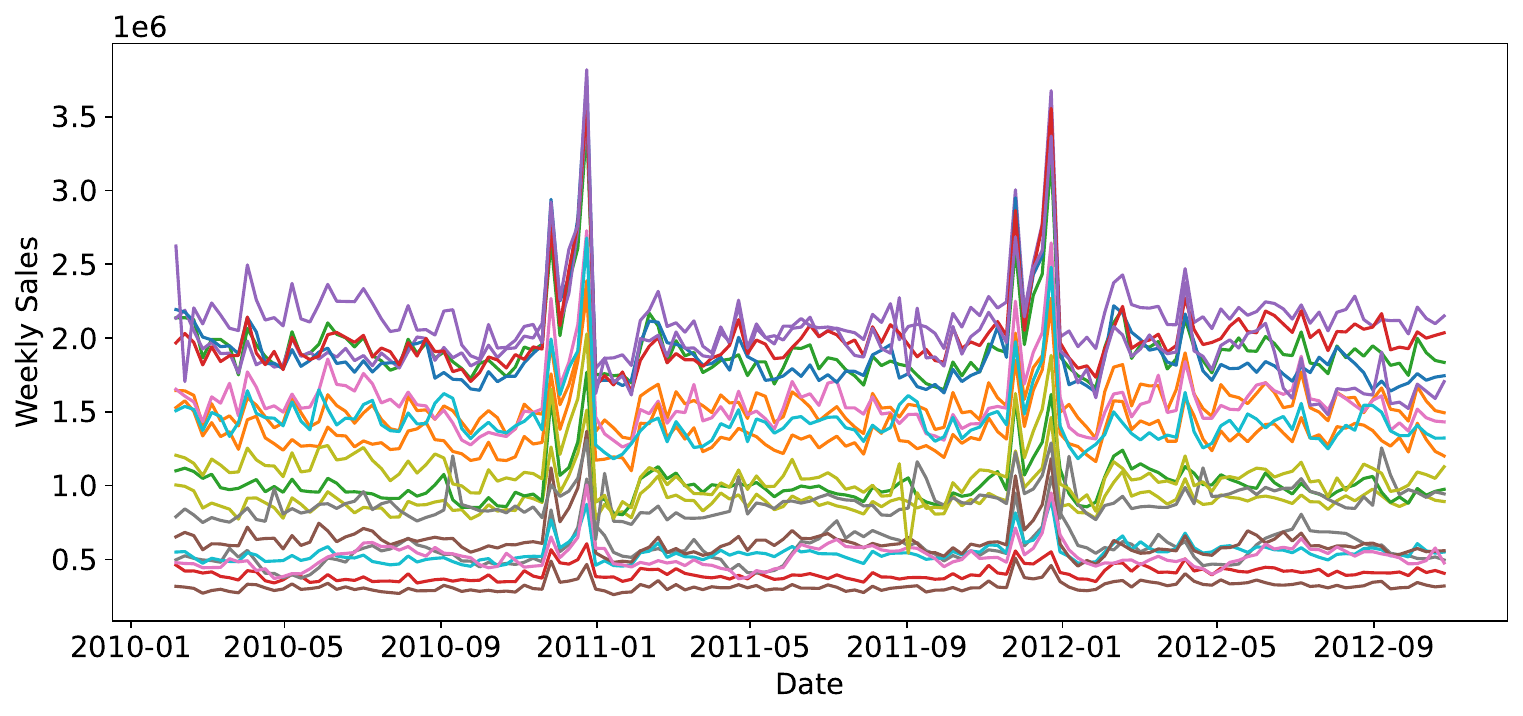}
\caption{Weekly sales of multiple stores in the Walmart data} 
\label{fig:shops45}
\end{center}
\end{figure}


\begin{enumerate}
    \item {\it Standalone Learning.} Each store in the dataset trains a model to forecast weekly demand without sharing information with other stores.
    \item {\it Centralized Learning.} All the stores share their data to a Centralized sever that trains a model on the aggregated global dataset.
    \item {\it Federated Learning.} The stores engage in a horizontal Blockchain FL mechanism to share model parameters, while keeping the dataset private
\end{enumerate}

To better compare these approaches, we adopt the same machine learning model in all the three cases:

\begin{itemize}
\item {\it Data preprocessing}. Numerical values are standardized using z-score, and categorical variables (weekday, month, year) are extracted from the \textit{Date} field.
\item {\it Train-test split}. Stores use either a 70-30 or 30-70 train-test ratio to simulate varying data availability and the benefits of data sharing.
\item {\it Machine Learning Model}. The forecasting was performed client-side using a simple neural network, with feature engineering achieved through lag methods. Regularization techniques such as dropout and weight decay were employed to prevent the model from overfitting. Leaky ReLU activation functions were used to maintain gradient flow during backpropagation, speeding up learning compared to the standard ReLU. More complex neural networks, such as LSTM or Attention-based models, were not used for forecasting to facilitate better comparison between the stand-alone, Centralized, and Federated versions. Particularly in the Federated case, an oversized network relative to the problem could lead clients to overfit on their specific data.
\end{itemize}

\subsection{Towards Trustworthy Federated Learning: Privacy and Transparency in Focus}
\label{exp:cyberattacks}

Our system prioritizes security and transparency among entities. To address this, we integrated several privacy and transparency mechanisms, as outlined in \cite{mazzocca2023framh}:

\textit{Global Differential Privacy} \cite{9069945} offers a balance between data privacy and model performance, mitigating multiple attack vectors such as \textit{Membership Inference Attack} and \textit{Reconstrution Attacks}. However, our experiments showed that the data is highly sensitive to noise. While local DP enhances privacy, it exponentially amplifies noise, degrading performance.

\textit{Secure Aggregation Plus} \cite{bonawitz2016practicalsecureaggregationFederated}: to enhance security, we implemented this technique, leveraging additive masking and secret key sharing among clients. This approach eliminates the need for Homomorphic Encryption, reducing computational overhead while improving resilience against threats such as \textit{Man-in-the-Middle} attacks and targeted backdoor injections.

For transparency, we integrated an Ethereum-based Blockchain \cite{Li2020}, enabling clients to verify server operations while ensuring data immutability and server transparency. This approach prevents manipulation, restricts malicious identity creation, and safeguards model integrity through IPFS storage.

\vspace*{0.07in}
\subsection{Experimental Setup}
\label{Experimental Setup}

\begin{figure}[!b]
\begin{center}
\includegraphics[width=8.8cm]{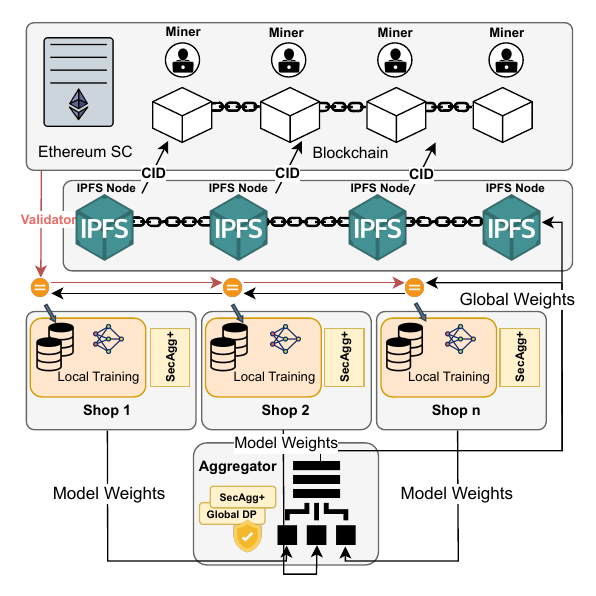}
\caption{Overview of the final architecture representing the correlation between the various entities and the lifecycle of the Federated Learning system, from local training to validation with Blockchain.} 
\label{fig:Architecture Overview}
\end{center}
\end{figure}

In our experimental setup (Fig. \ref{fig:Architecture Overview}), the architecture aggregates masked parameters via \textit{SecAgg+} \cite{bonawitz2016practicalsecureaggregationFederated} at the server level; gradients are clipped, and local training is performed using the \textit{FedAvg} Algorithm \ref{alg:fl_fedavg_secagg}. We also tested other aggregation algorithms, such as \textit{FedProx} and \textit{FedAdam}, but \textit{FedAvg} proved particularly suitable given the low heterogeneity of the data.

\begin{algorithm}
\small
\caption{Federated Learning with FedAvg, SecAgg+, and Global DP with Gradient Clipping}
\label{alg:fl_fedavg_secagg}
\begin{algorithmic}[1]
\State \textbf{Input:} $K$ clients, global model $M_0$, rounds $T$, clipping threshold $C$, noise $\epsilon$ for DP
\State \textbf{Output:} Global model $M_T$

\For{round $t = 1$ to $T$}
    \State Initialize $\Delta M = 0$, $n_{active} = 0$
    
    \For{each client $k = 1$ to $K$}
        \State Sample subset $S_t \subseteq \{1, \dots, K\}$
        
        \If{client $k \in S_t$}
            \State Compute local gradients $\nabla L_k$
            \State Clip gradients: $\nabla L_k \gets \min(C, \max(\nabla L_k, -C))$
            \State Add noise for DP: $\nabla L_k \gets \nabla L_k + \mathcal{N}(0, \sigma^2)$
            \State Send $\nabla L_k$ to server
            \State Increment $n_{active}$
        \EndIf
    \EndFor

    \State \textbf{Secure Aggregation (SecAgg+)}
    \State Aggregate gradients with SecAgg+: the server sees only the aggregated result
    \State Compute global update: $\Delta M_t = \frac{1}{n_{active}} \sum_{k \in S_t} \nabla L_k$

    \State \textbf{Model Update}
    \State Update global model: $M_{t+1} \gets M_t - \eta \Delta M_t$
\EndFor
\State \textbf{Return:} $M_T$
\end{algorithmic}
\end{algorithm}

\begin{algorithm}
\small
\caption{IPFS and Blockchain (Client-Server)}
\label{alg:ipfs_and_Blockchain}
\begin{algorithmic}[1]
\State \textbf{Server Side:}

\State \textbf{Step 1: Server stores CID on Blockchain}
\State Server sends model weights $M_k$ to IPFS
\State IPFS returns $\texttt{cid}_k$ for $M_k$
\State Server stores $\texttt{cid}_k$ on Blockchain

\State \textbf{Step 2: Server sends model update to clients}
\For{each client $k$}
    \State Server sends updated model $M_t$ to client $k$
\EndFor

\State \hrulefill

\State \textbf{Client Side:}

\State \textbf{Step 3: Client uploads model to IPFS}
\State Client sends model weights $M_k$ to IPFS
\State IPFS returns CID $\texttt{cid}_k$ for the uploaded model

\State \textbf{Step 4: Client calculates CID for updated model}
\State Client calculates CID $\texttt{cid}_k^{\prime}$ for the received model update $M_t$

\State \textbf{Step 5: CID verification}
\If{$\texttt{cid}_k^{\prime} = \texttt{cid}_k$ (from Blockchain)}
    \State Client confirms model update validity
\Else
    \State Client raises an alert for inconsistency
\EndIf
\end{algorithmic}
\end{algorithm}

\begin{figure*}[htb]
    \centering
    \subfloat[Mean Squared Error (MSE) of the three approaches considered (standalone, Federated, Centralized).]{
        \includegraphics[width=0.38\textwidth]{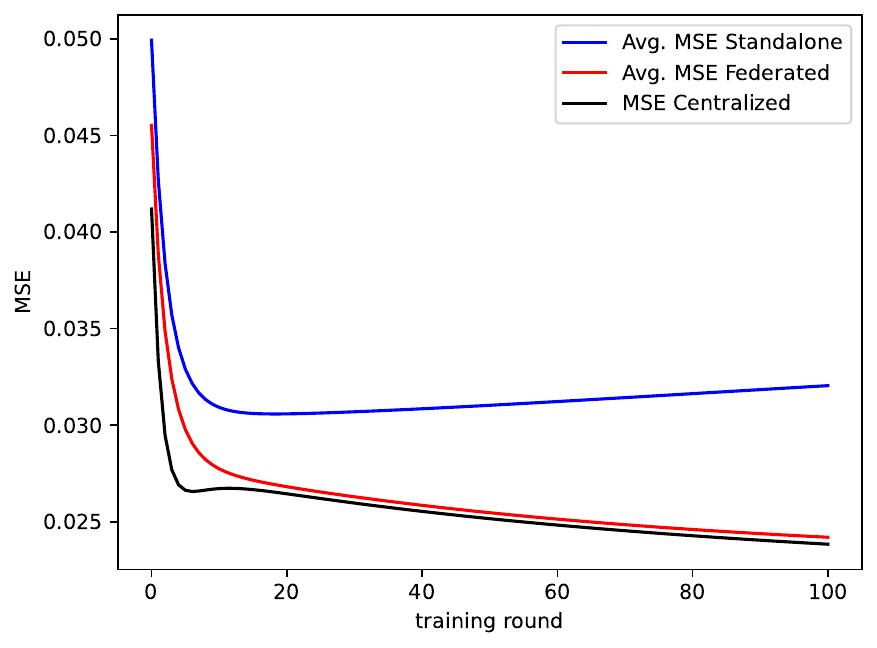}
    }
    \hfill
    \subfloat[Individual stores' MSE for Federated learning.]{
        \includegraphics[width=0.29\textwidth]{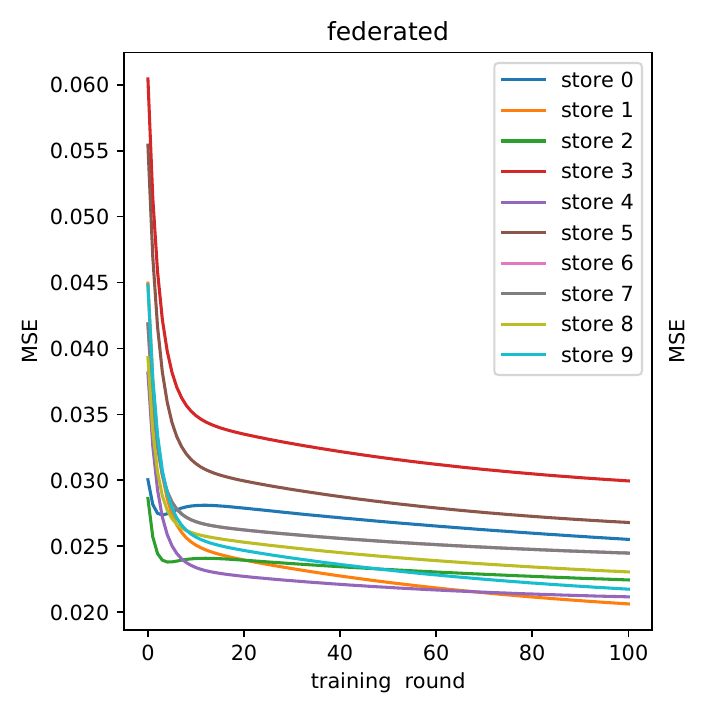}
    }
    \hfill
    \subfloat[Individual stores' MSE for standalone/Centralized learning.]{
        \includegraphics[width=0.27\textwidth]{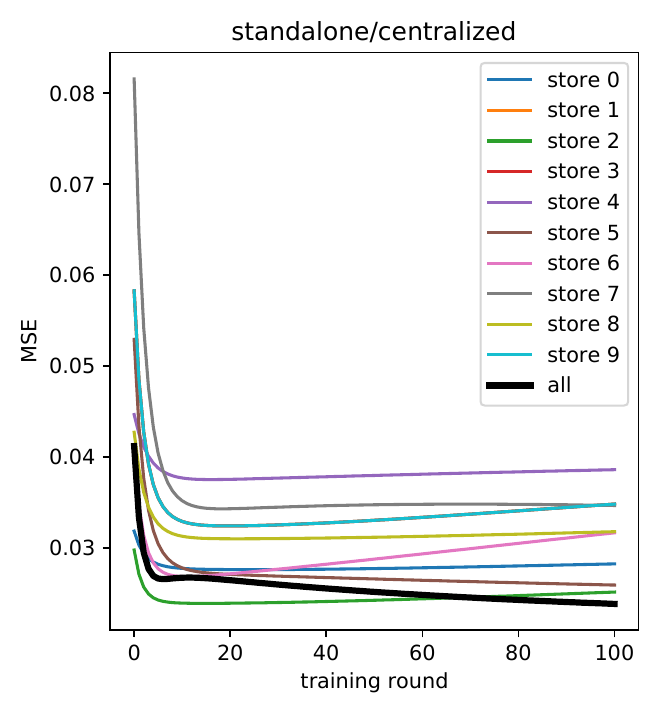}
    }
    \caption{Comparison of MSE for different learning approaches.}
    \label{fig:fl_vs_all_vs_single}
\end{figure*}

\begin{table*}[ht]
    \centering
    \caption{Blockchain comparison table}\label{cost comparison}
    \begin{tabular}{l c c c c c l}
\hline
Name & Policy & SC Deploy (Gwei) & Transaction cost (Gwei) & Type & Validation cost (Gwei) & Overall cost (ETH)\\
\hline
Ethereum & PoS & 2846556.504 & 93457.056 & Mainnet & 676405.576 & 0.070487114\\
Sepolia & PoS & 1569184.92 & 51518.88 & Testnet & 372873.48 & 0.038856533\\
Optimism & PoS & 688.239 & 22.596 & Layer 2 & 163.541 & 1.70423E-05\\
Arbitrum & PoS & 8947.107 & 293.748 & Layer 2 & 2126.033 & 0.00022155\\
Polygon PoS & PoS & 72953334 & 2395176 & Side-chain & 17335346 & 1.806487934\\
Polygon zkEVM & PoS & 6977366.982 & 229078.248 & Layer 2 & 1657978.658 & 0.172775233\\
\hline
\end{tabular}
\parbox{0.85\textwidth}{
    \footnotesize{Note: 1 Gwei ($10^{-9}$ ETH) - Prices based on the current rate as of 30/01/2025, 17:00 CET.}}
\end{table*}

Then Differential Privacy is applied to the global model, which, before being sent to the clients, is saved on IPFS as shown in Algorithm \ref{alg:ipfs_and_Blockchain}, a decentralized file system that allows the saving and compression of model weights into a CID in hash format, which is then stored on the Blockchain along with other relevant information, such as the timestamp and the number of rounds, ensuring that each transaction on the Blockchain is unique and inviolable. The use of IPFS is crucial because it helps optimize the computational load on a standard Blockchain that is highly secure but not suitable for high loads like Ethereum. It is important to highlight that, unlike some systems present in the state of the art, this solution is not fully decentralized, and the server remains the single point of failure. This choice was made deliberately because, as it is a horizontal Cross-Device Federated learning (entities belonging to the same organization), the level of trustability offered by this solution is already sufficiently high, and there is no need for a hierarchical structure between servers or to perform computationally expensive calculations on-chain.

\section{Experimental Evaluation}
\label{evaluation}


\subsection{ Sales Data Forecasting Across Different Scenarios }
The first analysis focuses on sales data forecasting across three distinct scenarios. As illustrated in Figure \ref{fig:fl_vs_all_vs_single}, the error in Centralized data handling is lower than in FL, which, in turn, performs better than standalone forecasting. Despite FL exhibiting slightly lower accuracy compared to Centralized data handling, its primary advantage lies in preserving data privacy. This characteristic is particularly valuable in real-world scenarios involving multiple companies, where data confidentiality is a critical concern.

\subsection{ Over-Provisioning Error and Waste Reduction in FL }

To further assess the effectiveness of Federated Learning (FL) in improving demand forecasting and reducing waste in grocery retail, we computed the over-provisioning error (OE). OE quantifies the extent to which forecasts exceed actual demand, a crucial factor in inventory management for perishable goods. Over-forecasting leads to excess stock, increasing waste, storage costs, and environmental impact. Given that food waste contributes significantly to global greenhouse gas emissions, reducing OE directly aligns with sustainability goals and supply chain efficiency.  

Our study evaluated OE across different forecasting approaches: standalone learning, Centralized learning, and FL. Standalone learning, where each store relies solely on its own sales history, often suffers from data sparsity, leading to inaccurate forecasts and sensitivity to biases. Centralized learning, which aggregates data across stores, improves predictive accuracy but raises concerns about data privacy and competitive confidentiality. FL, by contrast, enables collaborative model training without direct data sharing, mitigating these concerns while leveraging broader insights for improved forecasting.  

Results indicate that, on average, FL reduces OE, and consequently, perishable waste, by over 5\%. This reduction is particularly pronounced in stores where standalone forecasting performed poorly, with waste reductions exceeding 40\%. The effectiveness of FL in such cases underscores its ability to enhance forecasting for stores with limited data, improving inventory decisions and reducing inefficiencies. Furthermore, FL fosters a cooperative approach to demand forecasting, allowing retailers to benefit from shared intelligence without compromising sensitive data. As a result, FL presents a viable, privacy-preserving solution for optimizing inventory management and promoting sustainability in grocery retail.

\subsection{ Blockchain Platform Evaluation }
Additionally, we evaluated the computational costs associated with Blockchain integration by replicating the experiment on different Blockchain platforms. Five Ethereum-based Blockchains were tested, operating under the same smart contract (SC) conditions. The results indicate that while gas usage remains unchanged across platforms, gas prices vary significantly depending on network congestion and historical market conditions. As highlighted in Table \ref{cost comparison}, Layer 2 solutions offer significantly lower costs and greater scalability compared to Ethereum mainnet or its associated sidechains. However, these solutions may introduce security and interoperability concerns when interacting with non-Ethereum-based Blockchains. The trade-off between cost, performance, and security remains a key challenge in Blockchain-based FL.

Moreover, in more complex deployments involving a larger number of clients, token-based participation incentives, client reputation tracking, and the potential recording of partial updates, operational costs can increase up to fivefold. For instance, estimated costs range from approximately $8.50 \times 10^{-5}$ ETH for Optimism to 4.03 ETH for Polygon PoS, based on prevailing gas prices. Therefore, optimizing SC operations to minimize costs remains an open challenge for scalable and cost-effective Blockchain-based FL.F

\section{CONCLUSION}
\label{conclusions}

The adoption of FL in supply chain and manufacturing is often hindered by limited data access due to regulatory and market restrictions. Yet, collaborative learning models could benefit all stakeholders. This creates a prisoner's dilemma, where parties would gain from cooperation but hesitate due to data-sharing concerns. Privacy-preserving FL (PPFL), combined with security mechanisms, offers a viable solution by enabling model training without direct data exchange. Our results show that Blockchain-FL achieves performance close to the ideal Centralized scenario and significantly outperforms isolated models with a trade-off given by the necessary optimization of the smart contract in the Blockchain system to avoid excessive expenses in terms of gas fees.

Although FL has been successfully applied in retail, large-scale supply chains with frequent supplier changes present challenges. Encouraging participation from data-rich buyers may require incentive mechanisms, such as token-based rewards. Future work should explore broader use cases, addressing challenges like demand amplification (bullwhip effect) and shortage gaming. Establishing standardized communication protocols and regulatory frameworks will be crucial for enabling widespread adoption of FL in supply networks.

\section*{Acknowledgment}

This work has been supported by the project "A catalyst for EuropeaN ClOUd Services in the era of data spaces, high-performance and edge computing(NOUS)", Grant Agreement Number 101135927, and by the NextGenerationEU project: Ecosystem for Sustainable Transition in Emilia-Romagna (Ecosister) CUP: B33D21019790006 - PNRR - Missione 4 Componente 2 Investimento 1.5 - Spoke 4.

\bibliographystyle{IEEEtran}
\bibliography{ifacconf, Bib/Food_Waste, Bib/Sustainable_SCM, Bib/Federated_Learning, Bib/Information_Sharing}

\end{document}